\documentclass{article} 
\usepackage{iclr2027_conference,times}

\usepackage{amsmath,amsfonts,bm}

\def\eqref#1{equation~\ref{#1}}

\def\1{\bm{1}}

\DeclareMathAlphabet{\mathsfit}{\encodingdefault}{\sfdefault}{m}{sl}
\SetMathAlphabet{\mathsfit}{bold}{\encodingdefault}{\sfdefault}{bx}{n}

\usepackage[hidelinks]{hyperref}
\usepackage{url}
\usepackage{booktabs}
\usepackage{graphicx}
\usepackage{booktabs}
\usepackage{tabularx}
\usepackage{array}
\usepackage{makecell}
\newcommand{\method}{AWAM}
\hypersetup{
  pdftitle={AWAM: Active Trajectory-Conditioned View Imagination for Robust Robot Manipulation},
  pdfauthor={Jiawei Geng, Zhenyang Liu, Yuzheng Wu, Yanwei Fu},
  pdfkeywords={robot manipulation, active perception, world models, diffusion policies}
}

\title{\method{}: Active Trajectory-Conditioned View Imagination for Robust Robot Manipulation}

\author{Jiawei Geng$^{1}$, Zhenyang Liu$^{1,2}$, Yuzheng Wu$^{1}$, Yanwei Fu$^{1,2}$\\
\normalfont $^{1}$Fudan University\\
\normalfont $^{2}$Shanghai Innovation Institute}

\iclrfinalcopy
\begin{document}

\maketitle

\begin{abstract}
Rapid progress in embodied intelligence has enabled increasingly capable robots to follow language instructions and manipulate diverse objects. However, strong results in structured environments do not automatically transfer to open-world conditions, where clutter, occlusion, and changing viewpoints can obscure the spatial relationships required for action. To address this perceptual gap, a virtual active-perception paradigm is introduced. \method{}, an Active World-Action Model actively plans where to observe and imagines what will be seen along the planned trajectory before acting. Task-related virtual camera motion is predicted and used as a structured query into a geometry-conditioned world model, producing a reusable spatial prior for manipulation, which is later compressed into a fixed memory and fused with live observations through cross-attention for diffusion-based action prediction. \method{} achieves competitive success rates with 75.9\% on LIBERO-Occ, 62.4\% on LIBERO-Plus, 76.5\% on RLBench-OG Occlusion1, and 94.4\% in real-robot trials. These results support improved robustness to the evaluated occlusions and distribution shifts, addressing part of the gap between structured manipulation and less controlled environments.
\end{abstract}

\begin{figure}[h]
    \centering
    \includegraphics[width=\textwidth]{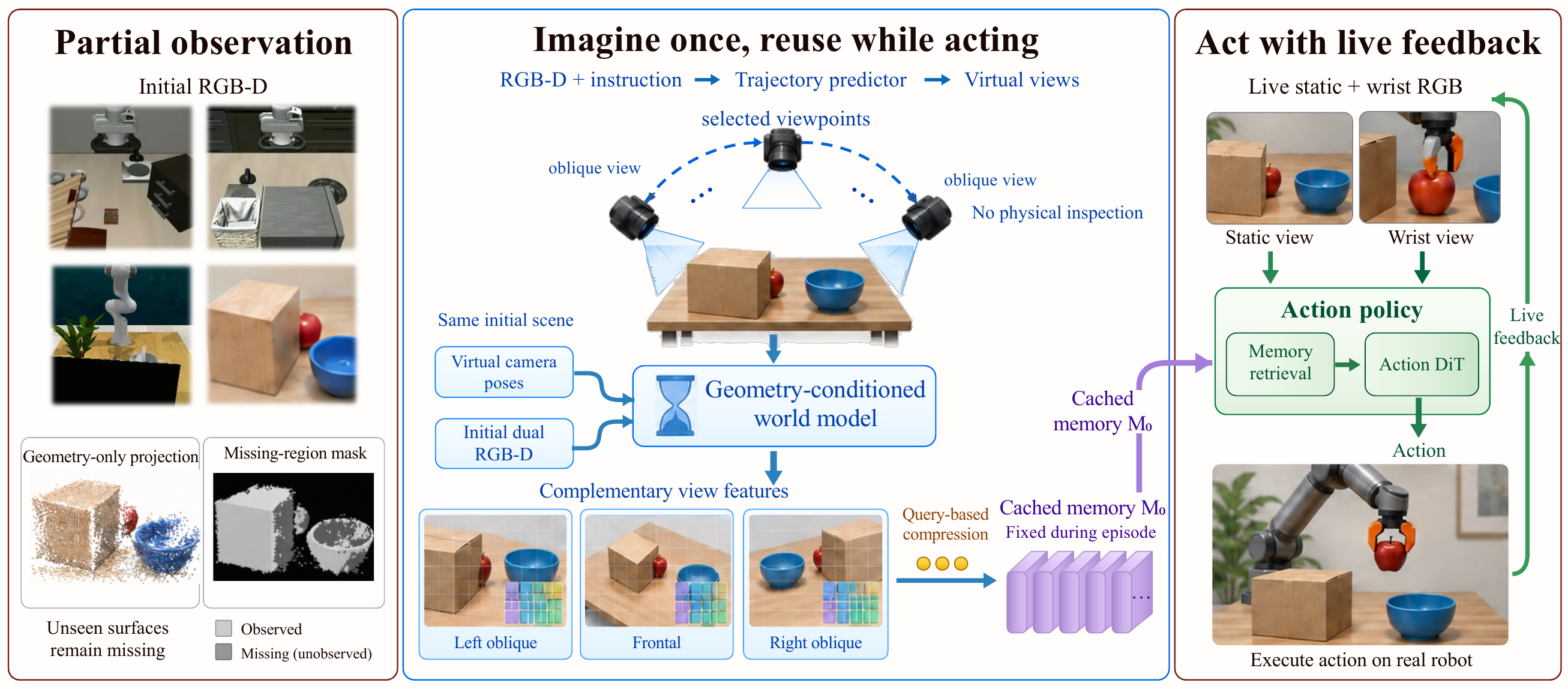}
    \caption{
    Overview of \method{}.
    From partially observed initial RGB-D inputs, \method{} imagines
    task-conditioned virtual views and compresses them into a predictive memory
    once per episode.
    The policy then reuses this memory together with live visual observations for
    closed-loop manipulation, without physically executing the predicted camera
    trajectory.
    }
    \label{fig:teaser}
\end{figure}

\section{Introduction}

Embodied intelligence has advanced rapidly through large-scale robot learning, vision--language pretraining, and generative action modeling. Increasingly capable policies can translate language goals and visual observations into manipulation behaviors across diverse tasks~\citep{rt1,rt2,octo,kim2024openvla,black2024pi0,chi2023diffusion}. 

However, much of this progress is evaluated in relatively structured settings in which camera configurations, scene layouts, and task-relevant objects remain sufficiently observable. Occlusion-focused and distribution-shift benchmarks expose a complementary limitation: once deployment becomes less structured, an unfavorable camera pose, clutter, or a shifted layout can hide the target or interaction region and remove the evidence required for control~\citep{li2026liberoocc,fei2026liberoplus,bai2026tvve,liu2025reasongrounder,liu2025neural}. This creates a perceptual bottleneck rather than only an action-modeling bottleneck. More expressive action models alone cannot directly recover scene evidence that is absent or highly ambiguous in the input, so performance gains obtained under well-observed scenes may not transfer to open-world conditions with incomplete visibility.

Several lines of work have begun to address this problem. Physical active perception acquires additional evidence by moving the camera or robot, but introduces extra execution steps and can alter the robot or scene state. Virtual-view methods avoid this cost by selecting or synthesizing complementary observations. Candidate-view approaches choose from predefined cameras~\citep{liu2026activevla,xu2026univiewvla}; TVVE predicts task-conditioned virtual poses and re-renders reconstructed geometry~\citep{bai2026tvve}; and world-action models such as Motus and Fast-WAM exploit predictive visual modeling to strengthen action representations~\citep{motus,fastwam}. These directions establish the value of additional visual context, yet the underlying limitation is not fully resolved: virtual perception is often restricted to isolated views, reconstruction-based rendering cannot reveal surfaces absent from the initial geometry, and predictive world models are commonly organized around action-induced futures rather than deliberate viewpoint queries.

To address these questions, task-conditioned virtual active perception is formulated as trajectory-conditioned view imagination. Instead of selecting a single view or a small set of independent poses, a virtual camera trajectory specifies how perception should evolve to expose task-relevant evidence. Such a trajectory can reorient toward a target, approach fine interaction geometry, or move around an occluder when no single viewpoint is sufficient. Importantly, the predicted trajectory is never physically executed; it serves only as a sequence of virtual queries, thus retrieving the available evidence and learned spatial priors into a representation that better supports control, separating perceptual exploration from robot motion.

Based on this formulation, \method{} is introduced as an Active World-Action Model that realizes the virtual-perception process before closed-loop control. A task-conditioned trajectory predictor first produces virtual camera poses for better observation from the initial RGB-D image and language instruction. The calibrated observations are then back-projected into coarse 3D geometry and rendered along the predicted path. A geometry-conditioned world model fills unsupported regions and produces multi-view latent features, which are compressed into a persistent world memory. During manipulation, live static- and wrist-camera observations retrieve relevant predictive context from this memory, and a diffusion policy generates short-horizon action chunks in closed loop. The resulting design uses world modeling as an internal perceptual mechanism for asking where to look and what would be seen, rather than as a prediction of the robot's executed future.

Evaluation is conducted across three simulation benchmarks and real-world manipulation. Success reaches 75.9\% on LIBERO-Occ, 62.4\% on LIBERO-Plus, $76.5\pm1.1$\% on RLBench-OG Occlusion1, and 94.4\% in real-robot trials. On LIBERO-Occ, removing the entire predictive-memory branch reduces average success by 11.5 percentage points, and replacing learned task-conditioned virtual trajectory with sampled viewpoints reduces it by 5.7 points. Together, these results demonstrate the effectiveness of learned virtual camera motion and persistent predictive memory for manipulation under incomplete visibility. The contributions are threefold:

\begin{itemize}

\item A trajectory-conditioned virtual active-perception paradigm is introduced for manipulation, replacing isolated or physically executed viewpoint acquisition with a continuous sequence of task-directed virtual queries.

\item \method{} is proposed to transform the predicted observation trajectory into geometry-consistent multi-view representations and compress them into persistent world memory as a reusable spatial prior for manipulation under partial observability.

\item Extensive evaluations across three challenging simulation benchmarks and real-world manipulation demonstrate strong performance under occlusion and distribution shifts and confirm the benefits of learned camera motion and predictive memory through ablations.

\end{itemize}

\section{Related Work}
\label{sec:related_work}
\paragraph{Robot learning and action generation.}
Large-scale robot policies map visual observations and language instructions directly to actions. Recent VLA models and diffusion-based policies improve generalization, action representation, and efficient decoding through large-scale pretraining and expressive generative objectives~\citep{rt1,rt2,octo,kim2024openvla,black2024pi0,chi2023diffusion,fast,openvla_oft,liu2025trivla,liu2025spatial}. These methods primarily focus on how actions are represented and learned from available observations. In contrast, \method{} targets the perceptual evidence supplied to the policy, constructing additional task-conditioned spatial context when relevant scene information is poorly observed or occluded.

\paragraph{Multiview manipulation under occlusion.}
Occlusion can be mitigated by acquiring or synthesizing complementary visual evidence through multiview perception, viewpoint selection, and imagined views~\citep{liu2026activevla,bai2026tvve,li2026liberoocc,xu2026univiewvla}. Existing approaches typically rely on additional sensing, reconstructed views, or a small fixed number of generated observations. \method{} instead predicts a task-conditioned virtual camera trajectory and consolidates the generated multi-view sequence into persistent world memory for closed-loop manipulation. This formulation turns occlusion handling into trajectory-conditioned predictive perception, where the model jointly reasons about where to observe, what would be observed, and how this information should support subsequent control.

\paragraph{World models for robotic manipulation.}
Recent video diffusion and world models have shown strong capability in modeling long-horizon visual dynamics, while text-, action-, and camera-conditioned variants further enable controllable prediction of future or novel observations~\citep{zhang2025gendop,zhang2026worldstereo,liu2026omnicam}. Predictive world models provide representations of how visual observations may evolve and have increasingly been integrated with robot control. Video Prediction Policy~\citep{hu2025vpp} conditions control on features learned from predictive video modeling. Motus~\citep{motus} unifies visual understanding, video generation, inverse dynamics, and action prediction within a latent world-action model. Fast-WAM~\citep{fastwam} separates the benefits of video co-training from explicit test-time future generation and shows that strong action prediction can be retained without iterative future-video rollout. Unlike approaches centered on predicting the future scene induced by robot actions, \method{} uses world modeling as a form of active perception. A task-conditioned camera trajectory specifies where the current scene should be queried, and a geometry-conditioned world model predicts visual representations along this trajectory. These representations are consolidated into a compact world memory and reused throughout manipulation rather than interpreted as the robot's executed future.

\section{Method}
\label{sec:method}

\begin{figure}[h]
\begin{center}
\includegraphics[width=\linewidth]{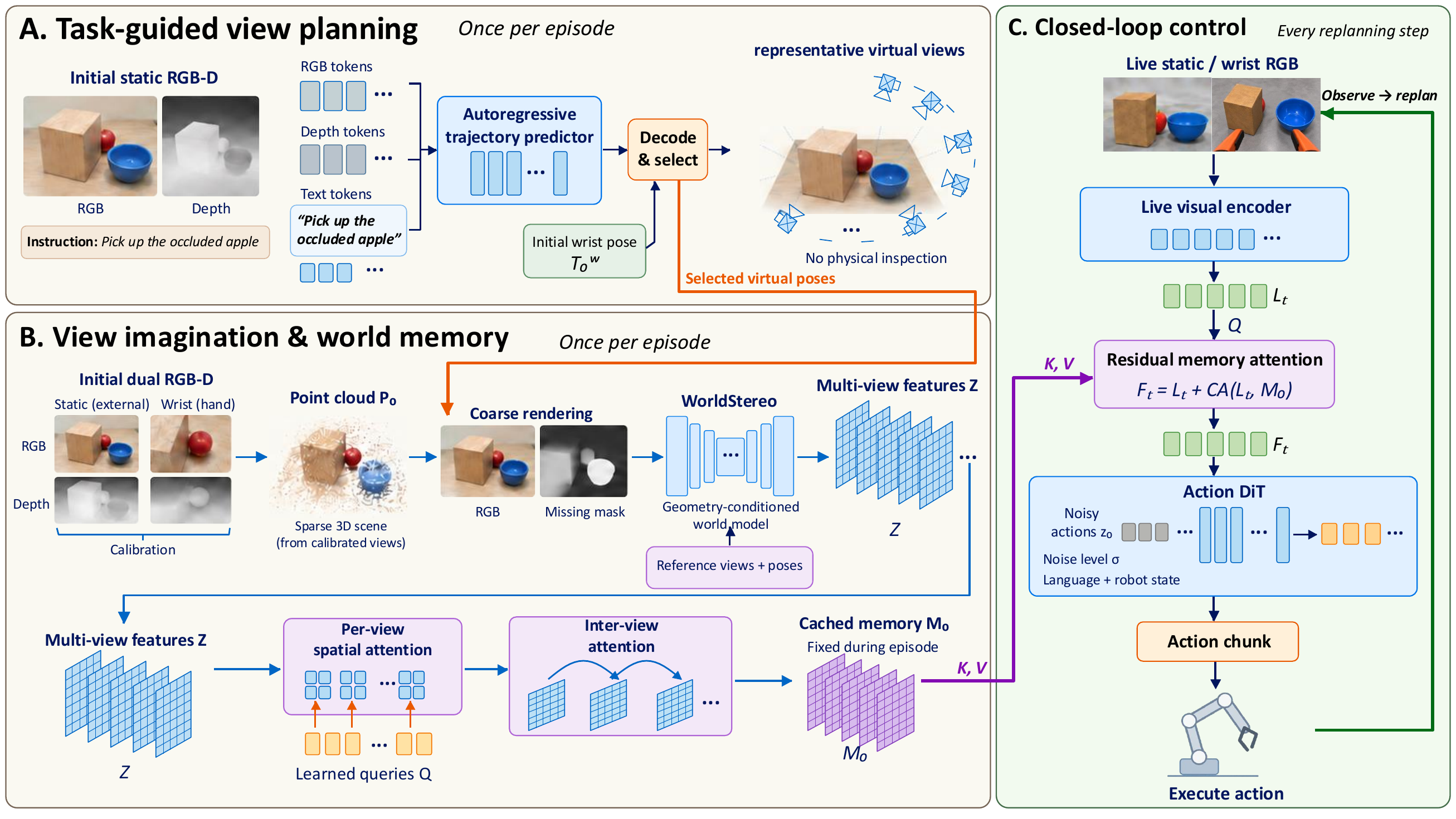}
\end{center}
\caption{Overview of \method{}. A task-guided virtual camera trajectory guides a geometry-conditioned world model to construct predictive memory once per episode. During closed-loop control, the cached world memory is fused with live visual observations to predict action chunks.}
\label{fig:pipeline}
\end{figure}

\subsection{Problem Formulation and Overview}
\label{sec:overview}


We consider language-conditioned manipulation with a static camera $s$ and a wrist-mounted camera $w$. At control step $t$, the policy observes RGB images $I_t^s$ and $I_t^w$, proprioceptive state $s_t$, and language instruction $\ell$.  At initialization it also receives metric depth and calibration: \begin{equation}
 \mathcal O_0=\left(\{I_0^v,D_0^v,K^v,T_0^v\}_{v\in\{s,w\}},s_0,\ell\right),
\end{equation}
where $K^v$ denotes camera intrinsics and $T_0^v\in\mathrm{SE}(3)$ maps camera coordinates to the world frame. Depth and calibration are used to build the initial memory; the action policy consumes live RGB observations and proprioception. \method{} separates memory construction from repeated action prediction (Fig.~\ref{fig:pipeline}): given the initial observation, \method{} constructs an active perception memory and subsequently conditions action prediction on both the memory and the current sensory evidence:
\begin{align}
 \widehat{\mathcal T}&=G_\phi(I_0^s,D_0^s,\ell;T_0^w), \label{eq:path-overview}\\
 M_0&=\mathcal R_\eta\left(
\mathcal F_\psi(\mathcal O_0,\widehat{\mathcal T})
\right), \label{eq:memory-overview}\\
 A_t&\sim\pi_\theta(\cdot\mid I_t^s,I_t^w,s_t,\ell,M_0).
\end{align}
Here, $G_\phi$ predicts task-relevant virtual viewpoints which may offer clearer observations, $\mathcal F_\psi$ extracts geometry-conditioned predictive representations along the planned perception trajectory, and $\mathcal R_\eta$ compresses them into a compact world memory. The action policy $\pi_\theta$ then combines this memory with live observations during closed-loop control.

\subsection{Task-Conditioned Virtual View Planning}
\label{sec:view-planning}

An unfavorable initial wrist-camera pose can hide the object needed for manipulation, leaving the policy to act on incomplete visual evidence. Rather than relying on the initial viewpoint alone, a task-conditioned camera trajectory is predicted to specify where additional virtual observations should be queried for better scene perception.

Demonstrated wrist-camera paths provide proxy supervision for task-relevant virtual queries. Each trajectory is parameterized relative to its initial wrist-camera pose, making prediction invariant to the global scene origin. For a wrist pose $T_i^w$, the relative transform is defined as $T_i=(T_0^w)^{-1}T_i^w$.

The relative poses are discretized into trajectory tokens encoding translation, rotation, and trajectory scale. Initial RGB-D observations and the language instruction are encoded into a shared conditioning representation $C$. A transformer then autoregressively predicts the trajectory sequence:
\begin{equation}
 \mathcal L_{\mathrm{traj}}=-\mathbb E_{\mathcal D}\sum_{j=1}^{J}
 \log p_\phi(y_j\mid C,y_{<j}).
\end{equation}
At inference, the decoded relative trajectory is transformed back into metric camera poses using the observed initial wrist pose $T_0^w$. A small set of representative viewpoints is retained along the trajectory rather than using every predicted pose. These virtual viewpoints define the camera queries passed to the world model and provide diverse observations of task-relevant regions while avoiding redundant predictions from nearby poses.

\subsection{Geometry-Conditioned World Memory}
\label{sec:world-memory}
A virtual view contains both geometrically supported pixels and regions that the initial cameras have not observed. We first unproject initial RGB-D observations into a colored point cloud,
\begin{equation}
\mathcal P_0=\bigcup_{v\in\{s,w\}}
\operatorname{BackProject}(I_0^v,D_0^v,K^v,T_0^v).
\end{equation}
Rendering $\mathcal P_0$ at the predicted camera poses preserves evidence from observed surfaces, but leaves holes where the reconstruction has no support. Such renderings become increasingly incomplete under viewpoint changes, since surfaces that are occluded or outside the initial fields of view have no geometric support, often resulting in large holes and unreliable visual features.

WorldStereo~\citep{zhang2026worldstereo} is employed as a geometry-conditioned world model and adapted to manipulation scenes. Intermediate geometry-conditioned features are extracted instead of passing generated RGB video to the action model, following the predictive-representation perspective of~\citet{hu2025vpp}. The generation target is an alternate view of the initial scene: objects and the robot are intended to remain fixed while the virtual camera changes pose.

The memory encoder compresses the predicted sequence. It aggregates spatial features within each predicted view and subsequently exchanges information across virtual viewpoints through inter-view attention. The resulting fixed-size token set summarizes the trajectory-level visual prediction while remaining efficient to retrieve during control, capturing what the robot is expected to observe from task-relevant  viewpoints while remaining grounded in the geometry of the current scene. 

\subsection{Memory-Conditioned Closed Loop Control}
\label{sec:action}

At each control step, the current static- and wrist-camera observations are encoded into live visual tokens $L_t$. Predictive context is retrieved from the cached world memory $M_0$ using residual cross-attention,
\begin{equation}
F_t =
L_t +
\operatorname{CA}(L_t,M_0),
\end{equation}
where $L_t$ serves as the attention query and $M_0$ provides the keys and values. Although the memory remains fixed throughout the episode, its retrieved content changes with the current observation. This allows the policy to access different parts of the predictive context as the manipulation state evolves, while the residual connection preserves a direct pathway from live perception to action prediction.

A conditional diffusion transformer is used to generate short-horizon action chunks. Given the fused visual representation $F_t$, proprioceptive state $s_t$, and language instruction $\ell$, the policy predicts
\begin{equation}
\widehat A_t=
\mathcal D_\theta
\left(
z_\sigma,\sigma;
F_t,s_t,\ell
\right),
\end{equation}
where $z_\sigma$ denotes the noisy action representation at diffusion level $\sigma$. The policy is trained on demonstrated action chunks using a standard diffusion denoising objective. During both action-policy training and evaluation, $M_0$ is constructed from the trajectory predicted by $G_\phi$ rather than from the demonstrated wrist-camera trajectory, avoiding a train--test mismatch in the memory input.

During closed-loop execution, the geometry-conditioned memory is constructed once from the initial observation and retained throughout the episode. At each replanning step, the policy receives fresh visual and proprioceptive observations, retrieves relevant predictive context, and generates a new action chunk from the updated state. The cached memory therefore acts as persistent predictive context about task-relevant virtual viewpoints rather than as a continually updated scene representation. Live observations remain the primary source of information about the evolving manipulation state, while the memory provides complementary spatial evidence that may be unavailable from the current viewpoint.

\section{Experiments}
\label{sec:experiments}
\subsection{Experimental Setup}
\label{sec:exp_setup}

\paragraph{Data and Environments.}
Evaluation is conducted on three complementary simulation benchmarks. LIBERO-Occ~\citep{li2026liberoocc} tests robustness to scene-induced occlusion across the Spatial, Object, Goal, and Long suites; all 2,000 benchmark instances are evaluated, with 500 rollouts per suite. The Occlusion1 split of RLBench-OG~\citep{bai2026tvve}, which contains ten occlusion-heavy manipulation tasks and uses a control interface different from LIBERO, provides a cross-benchmark test of the proposed world-memory mechanism. Both benchmarks directly test whether predicted viewpoints and imagined world memory predict task evidence omitted by the initial observation. Robustness beyond explicit occlusion is additionally evaluated on LIBERO-Plus~\citep{fei2026liberoplus}, which introduces controlled shifts in camera viewpoint, robot initial state, language instruction, illumination, background texture, sensor noise, and object layout. Environment-defined task success is reported for LIBERO-Occ and LIBERO-Plus, while mean $\pm$ standard deviation over three independent runs is reported for RLBench-OG. Real-world evaluation is conducted on an ARCHER-Y6 arm equipped with GP-100 grippers, scene-level and wrist-mounted Intel RealSense D435 cameras, and 100 original demonstrations.

\paragraph{Baselines.}
Baselines are selected according to the focus and evaluation ecosystem of each benchmark. For LIBERO-based evaluations, three categories are included: (i) representative vision-language-action policies such as $\pi_0$~\citep{black2024pi0}; (ii) world-model-augmented manipulation methods including Motus~\citep{motus} and Fast-WAM~\citep{fastwam}; and (iii) benchmark-specific occlusion methods represented by VIM~\citep{li2026liberoocc} on LIBERO-Occ. This set compares \method{} with both general-purpose policies and methods that explicitly introduce predictive or auxiliary visual representations. For RLBench-OG, the benchmark's evaluation ecosystem is followed using Diffusion Policy~\citep{chi2023diffusion}, ARP~\citep{arp}, RVT2~\citep{goyal2024rvt2}, and TVVE~\citep{bai2026tvve}. When released results do not cover the exact evaluation setting, the corresponding methods are reproduced and tuned using their recommended training recipes and evaluated under the same protocol as \method{}.



\subsection{Experimental Results}
\paragraph{Results on LIBERO-OCC.}
As shown in Table~\ref{tab:libero_occ}, \method{} achieves an average success rate of 75.9\%.
The improvement is consistent across three of the four suites: 84.6\% on Spatial, 93.2\% on Object, and 84.8\% on Goal. They benefit from obtaining task-relevant evidence that is difficult to infer from the current view: the predicted viewpoint sequence queries the initially hidden region, while the world model converts those queries into a persistent representation that can be reused during manipulation. On Long, however, Fast-WAM obtains the highest success rate (44.2\%), while \method{} reaches 41.0\%. Long-horizon tasks remain challenging because the cached memory is constructed from the initial scene and is not regenerated after substantial object-state changes, which might be solved by regenerating memory each time when gripper state changes. Nevertheless, \method{} attains the highest overall average, indicating that active observations remain useful across a broad range of occluded manipulation problems.

\begin{table}[t]
\caption{Success rate (\%) on LIBERO-Occ. Each suite contains 500 evaluation rollouts.}
\label{tab:libero_occ}
\centering
\small
\begin{tabular}{l|cccc|c}
\toprule
Method & Spatial & Object & Goal & Long & Avg. \\
\midrule
$\pi_0$  & 81.0 & 84.0 & 76.6 & 38.4 & 70.0 \\
Motus    & 82.2 & 78.0 & 66.0 & 43.4 & 67.4 \\
Fast-WAM & 82.8 & 86.4 & 78.8 & \textbf{44.2} & 73.0 \\
VIM      & 83.0 & 92.0 & 81.0 & 40.0 & 74.0 \\
\textbf{\method{}} & \textbf{84.6} & \textbf{93.2} & \textbf{84.8} & 41.0 & \textbf{75.9} \\
\bottomrule
\end{tabular}
\end{table}

\paragraph{Results on RLBench-OG.}
Table~\ref{tab:rlbench_og} presents the quantitative results on the Occlusion1 split of RLBench-OG.
\method{} achieves an average success rate of 76.5\%. 
The improvement is evident on tasks where the manipulation target or an interaction-relevant region is difficult to infer from the nominal observation.
For example, \method{} reaches 94.7\% success on water plants, improving over TVVE (88.0\%), ARP (66.0\%), and RVT2 (72.0\%), and obtains 92.0\% on scoop with spatula. It further achieves 100\% success on five other tasks.
Performance nevertheless remains low on block pyramid and solve puzzle.
Both require precise multi-stage interaction, for which recovering visibility alone is insufficient. This suggests that predictive observation is most useful when partial observability is a primary bottleneck, but does not eliminate the intrinsic control difficulty of high-precision long-horizon tasks. Nevertheless, the overall result and the favorable average rank in Table~\ref{tab:rlbench_og} show that the benefit of active world memory extends across a diverse set of occlusion-sensitive tasks and transfers beyond the LIBERO environment.

\begin{table*}[t]
\caption{
RLBench-OG Occlusion1 success (\%, mean $\pm$ std. over three runs). Avg. Rank is averaged across 10 tasks; lower is better.}
\label{tab:rlbench_og}
\centering
\scriptsize
\setlength{\tabcolsep}{4pt}
\renewcommand{\arraystretch}{1.10}

\begin{tabularx}{\textwidth}{
@{}
>{\raggedright\arraybackslash}p{0.16\textwidth}|
>{\centering\arraybackslash}X|
>{\centering\arraybackslash}X|
*{4}{>{\centering\arraybackslash}X}
@{}
}
\toprule
\textbf{Method}
&
\makecell{\textbf{Avg.} $\uparrow$}
&
\makecell{\textbf{Avg. Rank} $\downarrow$}
&
\makecell{basketball in\\hoop}
&
\makecell{scoop with\\spatula}
&
\makecell{take plate off\\dish rack}
&
\makecell{water\\plants}
\\
\midrule

Diffusion Policy
& $27.4{\pm}0.8$
& $4.70$
& $0.0{\pm}0.0$
& $0.0{\pm}0.0$
& $4.0{\pm}0.0$
& $2.0{\pm}2.0$
\\

ARP
& $73.0{\pm}1.6$
& $2.80$
& $\mathbf{100.0{\pm}0.0}$
& $88.0{\pm}4.0$
& $\mathbf{100.0{\pm}0.0}$
& $66.0{\pm}14.0$
\\

RVT2
& $72.8{\pm}2.0$
& $3.05$
& $\mathbf{100.0{\pm}0.0}$
& $82.7{\pm}7.5$
& $98.7{\pm}1.9$
& $72.0{\pm}8.6$
\\

TVVE
& $75.0{\pm}1.2$
& $2.40$
& $\mathbf{100.0{\pm}0.0}$
& $88.0{\pm}8.0$
& $\mathbf{100.0{\pm}0.0}$
& $88.0{\pm}4.0$
\\

\textbf{\method{}}
& $\mathbf{76.5{\pm}1.1}$
& $\mathbf{2.05}$
& $\mathbf{100.0{\pm}0.0}$
& $\mathbf{92.0{\pm}3.3}$
& $\mathbf{100.0{\pm}0.0}$
& $\mathbf{94.7{\pm}1.9}$
\\
\bottomrule
\end{tabularx}

\vspace{2pt}

\begin{tabularx}{\textwidth}{
@{}
>{\raggedright\arraybackslash}p{0.16\textwidth}
*{6}{>{\centering\arraybackslash}X}
@{}
}
\toprule
\textbf{Method}
&
\makecell{block\\pyramid}
&
\makecell{solve\\puzzle}
&
\makecell{take USB out\\of computer}
&
\makecell{close\\drawer}
&
\makecell{straighten\\rope}
&
\makecell{toilet seat\\down}
\\
\midrule

Diffusion Policy
& $0.0{\pm}0.0$
& $0.0{\pm}0.0$
& $92.0{\pm}8.0$
& $76.0{\pm}0.0$
& $0.0{\pm}0.0$
& $\mathbf{100.0{\pm}0.0}$
\\

ARP
& $\mathbf{14.0{\pm}2.0}$
& $0.0{\pm}0.0$
& $\mathbf{100.0{\pm}0.0}$
& $\mathbf{100.0{\pm}0.0}$
& $62.0{\pm}6.0$
& $\mathbf{100.0{\pm}0.0}$
\\

RVT2
& $4.0{\pm}3.3$
& $0.0{\pm}0.0$
& $\mathbf{100.0{\pm}0.0}$
& $\mathbf{100.0{\pm}0.0}$
& $\mathbf{70.7{\pm}10.0}$
& $\mathbf{100.0{\pm}0.0}$
\\

TVVE
& $8.0{\pm}8.0$
& $\mathbf{2.0{\pm}2.0}$
& $\mathbf{100.0{\pm}0.0}$
& $\mathbf{100.0{\pm}0.0}$
& $64.0{\pm}0.0$
& $\mathbf{100.0{\pm}0.0}$
\\

\textbf{\method{}}
& $12.0{\pm}3.3$
& $1.3{\pm}1.9$
& $\mathbf{100.0{\pm}0.0}$
& $\mathbf{100.0{\pm}0.0}$
& $65.3{\pm}3.8$
& $\mathbf{100.0{\pm}0.0}$
\\
\bottomrule
\end{tabularx}

\end{table*}

\paragraph{Results on LIBERO-PLUS.}
\label{sec:libero_plus}
As shown in Table~\ref{tab:libero_plus}, \method{} achieves an instance-weighted success rate of 62.4\%, outperforming $\pi_0$ (53.6\%), Fast-WAM (48.2\%), and Motus (44.0\%).
The gains are most pronounced under geometric perturbations that alter the observation configuration or scene structure, including camera viewpoint (53.4\%), robot-state (48.9\%), and object-layout shifts (73.0\%). Robustness to viewpoint changes is under known calibration. This pattern is consistent with the design of \method{}: instead of relying solely on the current observation, the policy can leverage task-conditioned visual evidence predicted from multiple geometrically related viewpoints, reducing its sensitivity to changes in camera pose and scene geometry.

\begin{table}[t]
\caption{Success rate (\%) on all 10,030 LIBERO-Plus instances; Total is instance-weighted.}
\label{tab:libero_plus}
\centering
\small
\begin{tabular}{l|ccccccc|c}
\toprule
Method & Camera & Robot & Language & Light & Background & Noise & Layout & Total \\
\midrule
$\pi_0$
& 13.8 & 6.0 & 58.8 & \textbf{85.0} & \textbf{81.4} & \textbf{79.0} & 68.9 & 53.6 \\
Motus
& 17.9 & 43.3 & \textbf{74.1} & 54.7 & 44.9 & 26.2 & 52.0 & 44.0 \\
Fast-WAM
& 18.8 & 44.5 & 68.9 & 78.2 & 45.7 & 29.8 & 60.7 & 48.2 \\
\textbf{\method{}}
& \textbf{53.4} & \textbf{48.9} & 61.0 & 84.3 & 61.1 & 60.7 & \textbf{73.0} & \textbf{62.4} \\
\bottomrule
\end{tabular}
\end{table}

\paragraph{Real-world evaluation.}

\method{} is evaluated on three real-world manipulation tasks in which the initial observations provide incomplete task-relevant information:
(i) retrieving a partially occluded apple from a box;
(ii) picking a bottle from a cluttered tabletop and placing it into a nearby box;
(iii) grasping a block initially occluded by a box and placing it into the box.

\begin{figure}[h]
\centering
\includegraphics[width=\linewidth]{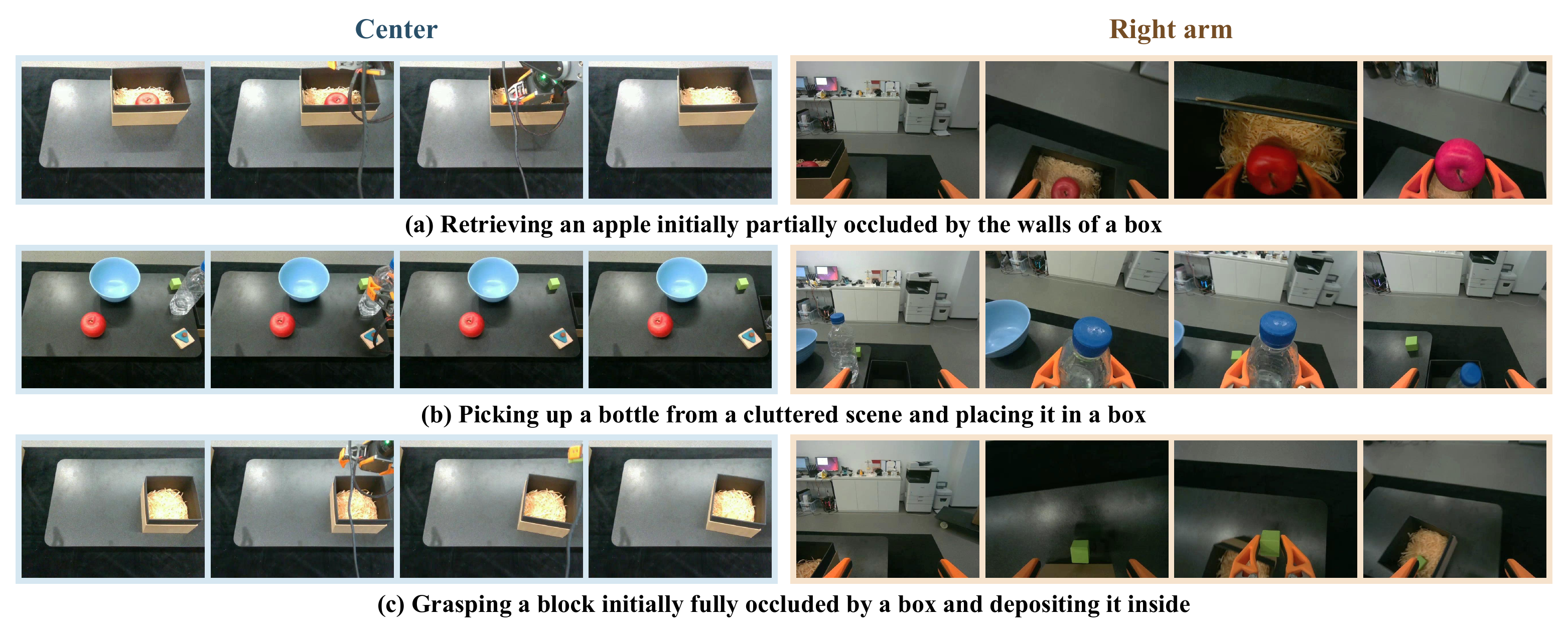}
\caption{Real-world rollouts of \method{} under partial occlusion and clutters}
\label{fig:real_robot}
\end{figure}

As illustrated in Fig.~\ref{fig:real_robot}, \method{} successfully handles partially occluded objects, cluttered scenes, and initially
hidden targets in real-world grasping and placement.
These rollouts highlight the practical value in addressing limited visibility and spatial ambiguity. 
Successful execution across these tasks indicates that the active-perception memory framework extends beyond simulation to real-world robotic settings.

The robot platform and camera configuration described in Sec.~\ref{sec:exp_setup} are used, with 30 trials per task for each method. As shown in Table~\ref{tab:real_robot}, \method{} achieves 85/90 successful trials (94.4\%), compared with 66/90 (73.3\%) for $\pi_0$, 73/90 (81.1\%) for Motus, and 72/90 (80.0\%) for Fast-WAM. 
The consistent gains under different occlusion configurations suggest that imagining task-conditioned virtual viewpoints and caching them as predictive memory can provide complementary visual evidence for closed-loop manipulation without requiring additional physical camera motion.

\begin{table}[t]
\caption{Real-world manipulation, successes/trials (success rate, \%).}
\label{tab:real_robot}
\centering
\small
\begin{tabular}{lcccc}
\toprule
Method &  apple in box & bottle in cluster & block behind box & Overall \\
\midrule
$\pi_0$ & 22/30(73.3\%) & 24/30(80.0\%) & 20/30(66.7\%) & 66/90(73.3\%) \\
Motus & 24/30(80.0\%) & 26/30(86.7\%) & 23/30(76.7\%) & 73/90(81.1\%) \\
Fast-WAM & 24/30(80.0\%) & 25/30(83.3\%) & 23/30(76.7\%) &  72/90(80.0\%)\\
\textbf{\method{}} & \textbf{28/30 (93.3\%)} & \textbf{30/30 (100.0\%)} & \textbf{27/30 (90.0\%)} & \textbf{85/90 (94.4\%)} \\
\bottomrule
\end{tabular}
\end{table}

\subsection{Ablation Study And Qualitative Analysis}

\paragraph{Component Analysis.}

To disentangle the contributions of each component, four controlled variants are evaluated while keeping the policy backbone, training data, and evaluation
protocol unchanged. Results are summarized in
Table~\ref{tab:component_ablation}.
\begin{table}[t]
\centering
\small
\caption{
  Component analysis on LIBERO-Occ.
  $\Delta$ denotes the change from full \method{} in percentage points.
}
\label{tab:component_ablation}
\setlength{\tabcolsep}{4pt}
\begin{tabular}{lccccc}
\toprule
Variant & Learned trajectory & Generation & Memory & Success (\%) $\uparrow$ & $\Delta$ \\
\midrule
(a)Live observations only & $\times$ & $\times$ & $\times$ & 64.4 & -11.5\\
(b)Trajectory-conditioned live policy & \checkmark & $\times$ & $\times$ & 67.3 & -8.6 \\
(c)sampled spherical views  & $\times$ & \checkmark & \checkmark & 70.2 & -5.7 \\
(d)Learned path, warp only  & \checkmark & $\times$ & \checkmark & 69.0 & -6.9 \\
\midrule
\textbf{\method{}} & \checkmark & \checkmark & \checkmark & \textbf{75.9} & -- \\
\bottomrule
\end{tabular}
\end{table}

Variant(a) shows that removing the entire predictive branch and relying only on live observations reduces the success rate from 75.9\% to 64.4\%, corresponding to the largest drop of 11.5 percentage points. This result confirms that the performance gain cannot be explained by the closed-loop policy alone, and that information constructed before control provides important complementary visual context when the current observations are incomplete or occluded. 
Variant(b), where the predicted camera trajectory is directly provided to the live policy without virtual-view generation or memory construction, performance improves to 67.3\%. The 2.9-point gain over the live-only variant suggests that the task-conditioned trajectory itself carries a useful implicit manipulation prior. However, its substantial gap to the full model indicates that trajectory conditioning alone cannot substitute for explicit predictive visual context.
Variant(c), which replaces the learned trajectory with spherical viewpoints uniformly sampled around the whole scene region achieves 70.2\%. Despite preserving view diversity as well as the same generation and memory modules, the 5.7-point degradation shows that generic coverage of the scene is less effective than allocating virtual observations according to a learned, task-conditioned trajectory. 
Variant(d) retaining the learned trajectory while removing generative prediction and constructing memory only from warped geometric observations yields 69.0\%. This variant preserves the views queried by the model, but can only reproject evidence already visible in the initial RGB-D observations. Its 6.9-point drop indicates that viewpoint selection alone is insufficient under severe occlusion, where previously unobserved surfaces cannot be recovered through geometric warping.

Taken together, these ablations reveal complementary roles of the proposed components. The predicted trajectory provides both task-dependent viewpoint selection and a weak planning prior, while generative prediction converts these queries into predictive visual context beyond the initially observed geometry. Compressing this predicted evidence into memory further allows the closed-loop policy to access such information throughout manipulation. The full combination therefore achieves the highest success rate of 75.9\% on LIBERO-Occ.

\paragraph{Sensitivity to Virtual-View Budget.}
The trajectory length $K$ determines the amount of virtual visual context constructed before control. $K$ varies in $\{7,14,21,28\}$ while keeping all other settings fixed. Performance improves as the trajectory provides more virtual observations and saturates around $K=21$, with success rates of 72.4\%, 74.8\%, 75.9\% and 76.0\% on LIBERO-Occ respectively. Increasing the length further to 28 brings no meaningful improvement, suggesting limited sensitivity to the exact choice of $K$ once sufficient task-relevant scene coverage is obtained.

\begin{figure}[h]
\centering
\includegraphics[width=\linewidth]{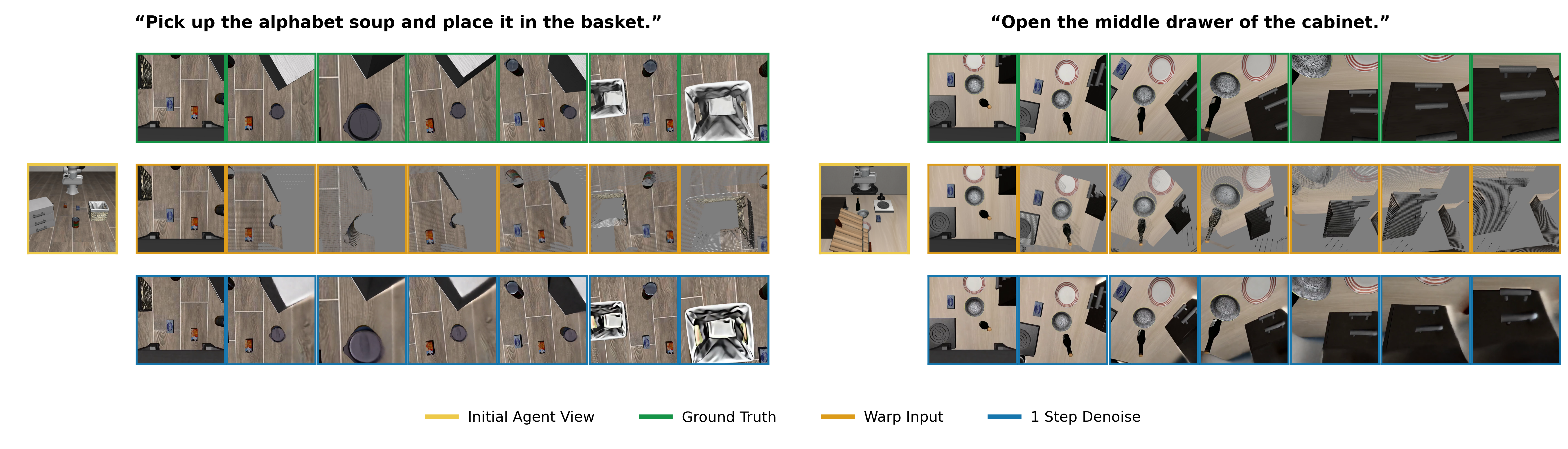}
\caption{Virtual-view predictions along the planned trajectory. Warped inputs (orange), predictions (blue), and ground truth (green) show progressively improved visibility; RGB decodes are diagnostic only.}
\label{fig:ws}
\end{figure}

\paragraph{Predicted Virtual Views.}
For policy conditioning, intermediate predictive features are extracted after the first denoising update rather than completing the full RGB generation process. The RGB decodes in Fig.~4 are used only for qualitative diagnosis. Since \method{} uses a predicted camera trajectory to construct task-conditioned world memory, observations generated along this trajectory are visualized in Fig.~\ref{fig:ws}. 

The initial agent view is often occluded and does not clearly expose the task-relevant object or interaction region.
As the predicted trajectory proceeds, the virtual camera progressively shifts toward more informative regions, revealing the target object, receptacle, or manipulation interface from more favorable perspectives. 
For instance, the virtual trajectory gradually exposes both the alphabet soup and the basket in the placement task, and moves toward the cabinet to reveal the middle drawer in the drawer-opening task. 

The warped inputs provide a coarse geometric projection under the predicted viewpoints, while a single denoising step substantially restores appearance and scene structure necessary for manipulation. The average hole rate of LIBERO-Occ is 39.29\%, with peak values reaching 74.47\%. Despite these large holes, the denoised predictions still recover the key object layouts and interaction regions, indicating that the trajectory predictor successfully identifies where to observe and that the world model can translate these virtual viewpoints into task-relevant visual evidence for downstream action generation.

\section{Conclusion}

Rather than relying only on the currently visible camera frames, \method{} predicts a task-conditioned virtual camera trajectory and uses a geometry-conditioned world model to imagine observations along this trajectory. The resulting representations are compressed into a persistent world memory and jointly used with live observations during closed-loop action prediction. Experiments across occlusion-focused and distribution-shift benchmarks show that this additional predictive context improves manipulation robustness when task-relevant objects or interaction regions are poorly observed. These results suggest that world models can serve not only as predictors of future dynamics, but also as internal perceptual mechanisms for querying spatial information that is missing from the current observation. Future work could selectively refresh memory when large state changes are detected, extending virtual active perception to longer-horizon and more dynamic manipulation.

\clearpage
\section*{AI use statement}
Generative AI tools were used to assist with programming, including code drafting, debugging, and implementation support, and to improve manuscript clarity, grammar, and presentation. They were not used to autonomously generate experimental results or determine the scientific conclusions of this work. All AI-assisted code was reviewed, tested, and validated by the authors before use in experiments, and AI-assisted manuscript edits were manually reviewed for technical accuracy, consistency with the underlying experiments, and appropriate attribution. The authors take full responsibility for the final content, including text, claims, code, and artifacts produced with the aid of generative AI.

\section*{Ethics statement}
This work does not involve human subjects, sensitive personal data, or applications with foreseeable societal risks. Experiments are conducted on publicly available datasets and widely used
pretrained models under their respective licenses. 

\section*{Reproducibility statement}
Multiple steps are taken to support reproducibility. The model architectures, training data and objectives, and evaluation protocols are specified in the paper, with additional implementation details provided in the appendix. The code will be released upon acceptance.

\bibliography{iclr2027_conference}
\bibliographystyle{iclr2027_conference}

\newpage
\appendix
\section{Appendix}
\subsection{Additional Implementation Details}
\label{app:implementation}

\subsubsection{Camera Representation and Virtual Trajectory}
\label{app:trajectory}

\paragraph{Coordinate convention.}
The static camera and wrist camera are denoted by $s$ and $w$, respectively.
A camera pose
$
T_i^w=[R_i\mid p_i]\in\mathrm{SE}(3)
$
maps camera coordinates to the world frame.
All poses used by the geometry pipeline follow the OpenCV camera convention,
with $x$ pointing right, $y$ down, and $z$ forward.
Metric depth and camera translation use the same physical unit.

As described in ~\ref{sec:view-planning}, the trajectory predictor operates on poses relative
to the initial wrist-camera pose:
\begin{equation}
    \overline T_i=(T_0^w)^{-1}T_i^w,
    \qquad
    T_i^w=T_0^w\overline T_i.
    \label{eq:app_relative_pose}
\end{equation}
This representation removes dependence on the absolute world-frame origin.
The world-model renderer consumes the corresponding world-to-camera
extrinsics obtained by rigid inversion.

\paragraph{Trajectory representation.}
Demonstrated wrist-camera trajectories are first temporally resampled.
Each relative pose is represented by rotation, normalized translation,
camera-related auxiliary coordinates, and a trajectory-scale variable.
The normalized coordinates are discretized into trajectory tokens and modeled
autoregressively.

For a normalized scalar $x\in[-1,1]$, the reference implementation uses
\begin{equation}
q(x)=
\left\lfloor
\operatorname{clip}
\left(
\frac{B(x+1)}{2},
0,
B
\right)
\right\rfloor ,
\qquad B=256 ,
\end{equation}
which gives integer values from $0$ to $256$.
The decoded trajectory is converted back to metric scale and anchored to
the observed initial wrist-camera pose using
Eq.~\ref{eq:app_relative_pose}.

\paragraph{Trajectory predictor.}
The trajectory generator is initialized from a text--RGB-D camera-trajectory
model. The static-camera RGB image, metric depth map, and language instruction
are independently encoded and projected into a shared conditioning sequence.
A causal transformer autoregressively predicts the trajectory tokens using
the teacher-forced objective of the main paper.

The visual and language backbone encoders remain frozen during trajectory
adaptation, while the trajectory-specific projection layers and causal
transformer are optimized for the manipulation-domain trajectories.

\paragraph{Selecting virtual viewpoints.}
Rather than sending every predicted pose to the world model, $K=21$ representative poses are retained along the trajectory.
To avoid allocating most of the view budget to nearly stationary trajectory
segments, viewpoints are sampled according to accumulated camera motion.

For two consecutive poses, the translation and rotation increments are computed as
\begin{align}
d_i &= \|p_i-p_{i-1}\|_2,\\
\vartheta_i &=
\arccos\left[
\operatorname{clip}
\left(
\frac{
\operatorname{tr}(R_{i-1}^{\mathsf T}R_i)-1
}{2},
-1,1
\right)
\right].
\end{align}
Translation and rotation increments are independently normalized by their
trajectory totals before being combined.
The selected poses are approximately uniformly distributed over accumulated
camera motion and include the trajectory endpoints.

Importantly, these poses are \emph{virtual queries}.
The robot never physically executes the predicted camera trajectory before
manipulation.

\subsubsection{Geometry-Conditioned Predictive Memory}
\label{app:memory}

\paragraph{Initial geometric evidence.}
At episode initialization, the calibrated static- and wrist-camera RGB-D
observations are back-projected into a joint colored point cloud.
For camera $v\in\{s,w\}$ and pixel
$\widetilde u=(u_x,u_y,1)^{\mathsf T}$ with valid depth $D_0^v(u)$,
\begin{equation}
X_0^v(u)
=
R_0^v
\left(
D_0^v(u)(K^v)^{-1}\widetilde u
\right)
+p_0^v .
\end{equation}
The two point sets are merged into the initial geometry
$\mathcal P_0$.

Rendering $\mathcal P_0$ from a predicted viewpoint preserves appearance
for surfaces supported by the initial observations, but leaves holes where
the initial cameras contain no geometric evidence.
This distinction motivates the predictive world model: point-cloud rendering
can rearrange observed surfaces, but cannot reconstruct surfaces that were
never observed.

\paragraph{Geometry-conditioned world model.}
WorldStereo~\citep{zhang2026worldstereo} is used as the predictive visual backbone and adapted to manipulation scenes.
The world model receives the coarse geometric rendering, missing-region mask,
camera information, and reference observations, and predicts
view-consistent representations along the virtual camera trajectory. The generation targets are static multi-view observations of the episode-initial scene. 
Supervision is obtained by holding the episode-initial simulator state fixed and
rendering additional calibrated camera poses without advancing the environment dynamics. Thus, differences across target views are caused by camera motion rather than object or robot motion.

The generation target corresponds to alternative observations of the initial scene.
The robot and scene objects are treated as stationary while the virtual
camera changes pose.
Thus, the predicted sequence should be interpreted as viewpoint-conditioned
scene imagination rather than an action-induced future rollout.

The policy consumes intermediate world-model features instead of generated RGB frames. RGB decoding is used only for the qualitative visualizations in Fig.~\ref{fig:ws} of the main paper.

\paragraph{Predictive feature extraction.}
For each of the $K=21$ virtual views, intermediate world-model features are
spatially pooled to a $16\times16$ grid.
In the reference implementation, this produces
\begin{equation}
Z\in\mathbb{R}^{21\times256\times5120}.
\end{equation}
The features remain separated by viewpoint during extraction.

\paragraph{Memory compression.}
The predictive features from the virtual trajectory contain dense spatial information across multiple viewpoints. Rather than exposing all features directly to the action policy, they are compressed with a lightweight query-based memory encoder.

For each virtual view, a small set of learned queries first aggregates its spatial features through cross-attention. Queries occupying the same slot then exchange information across the virtual-view sequence through inter-view attention. 

The outputs from all virtual views are flattened into a compact memory $M_0$, which is cached for the duration of the episode. This representation preserves trajectory-level predictive context. Detailed architectural dimensions of the reference implementation are reported in Table~\ref{tab:app_architecture}.

\subsubsection{Memory-Conditioned Closed-Loop Policy}
\label{app:policy}

\paragraph{Live visual encoding.}
At each replanning step, the latest static- and wrist-camera RGB observations are encoded by a shared visual backbone. Camera-specific embeddings preserve the identity of the two visual streams, after which a language-conditioned
resampler converts the spatial features into a fixed-size set of live visual tokens $L_t$.

Unlike the predictive memory, which is constructed once from the initial observation, $L_t$ is recomputed from fresh observations during execution. It therefore provides the policy with up-to-date information about the evolving manipulation state.

\paragraph{Memory retrieval.}
Predictive information is retrieved from the cached memory using residual cross-attention, where the live representation provides the attention queries and the cached memory provides keys and values.

Although $M_0$ remains fixed during an episode, the attention weights depend on the current observation. Consequently, different predictive information can be retrieved as the manipulation state evolves. The residual pathway
also preserves direct access to the current visual evidence.

\subsection{Reference Architecture Settings}
\label{app:architecture}

Table~\ref{tab:app_architecture} summarizes the architectural dimensions of the reference LIBERO implementation. These values are reported for reproducibility and should be interpreted as implementation settings rather than separate methodological contributions.

\begin{table}[t]
\centering
\small
\caption{
Reference architectural settings used in the LIBERO implementation of
\method{}. These values specify the implementation used in our experiments
and are reported for reproducibility.
}
\label{tab:app_architecture}
\setlength{\tabcolsep}{5pt}
\begin{tabularx}{\linewidth}{
@{}>{\raggedright\arraybackslash}X
>{\raggedright\arraybackslash}p{0.42\linewidth}@{}}
\toprule
Component & Setting \\
\midrule
Virtual-view queries & 21 viewpoints \\
World feature resolution & $16\times16$ tokens per view \\
World feature width & 5120 \\
Memory feature width & 384 \\
Memory encoder & 2 layers, 8 attention heads \\
Compressed memory & 294 tokens \\
Live visual representation & 224 tokens, width 384 \\
Memory retrieval & 8-head cross-attention \\
Memory lifetime & One episode \\
LIBERO action horizon & 8 actions \\
\bottomrule
\end{tabularx}
\end{table}

\subsection{Training Procedure}
\label{app:training}

The modules are optimized sequentially rather than through end-to-end backpropagation. The trajectory predictor is trained first, followed by world-model adaptation and then the action-policy stages. 

\paragraph{Trajectory model.}
The trajectory model is initialized from a pretrained text--RGB-D camera
trajectory generator and adapted using demonstrated wrist-camera paths.
The visual and language backbone encoders remain frozen, while the
trajectory-specific projection and autoregressive prediction layers are
optimized with the teacher-forced objective described in the main paper.

\paragraph{World model.}
The geometry-conditioned world model~\citep{zhang2026worldstereo} is adapted to manipulation scenes while retaining its pretrained geometric and camera-conditioning capabilities. Training targets correspond to alternative observations of the same initial scene under different camera poses. Thus, the model learns to predict viewpoint-conditioned scene representations rather than robot-induced future states.

During action-policy training, world-model features are precomputed and cached. This avoids repeatedly running the large predictive model during policy optimization and provides a fixed representation interface between predictive perception and action learning.

\paragraph{Action policy.}
Training is divided into an interface-alignment stage and an action-model adaptation stage.
The first stage optimizes the live visual projection, resampler, memory encoder, and memory-fusion modules while keeping the pretrained action denoiser fixed.
The second stage freezes these connection modules and adapts the diffusion action model.

\subsection{Additional Evaluation Details}
\label{app:evaluation}
\subsubsection{LIBERO-Occ}

\begin{figure}[h]
\centering
\includegraphics[width=\linewidth]{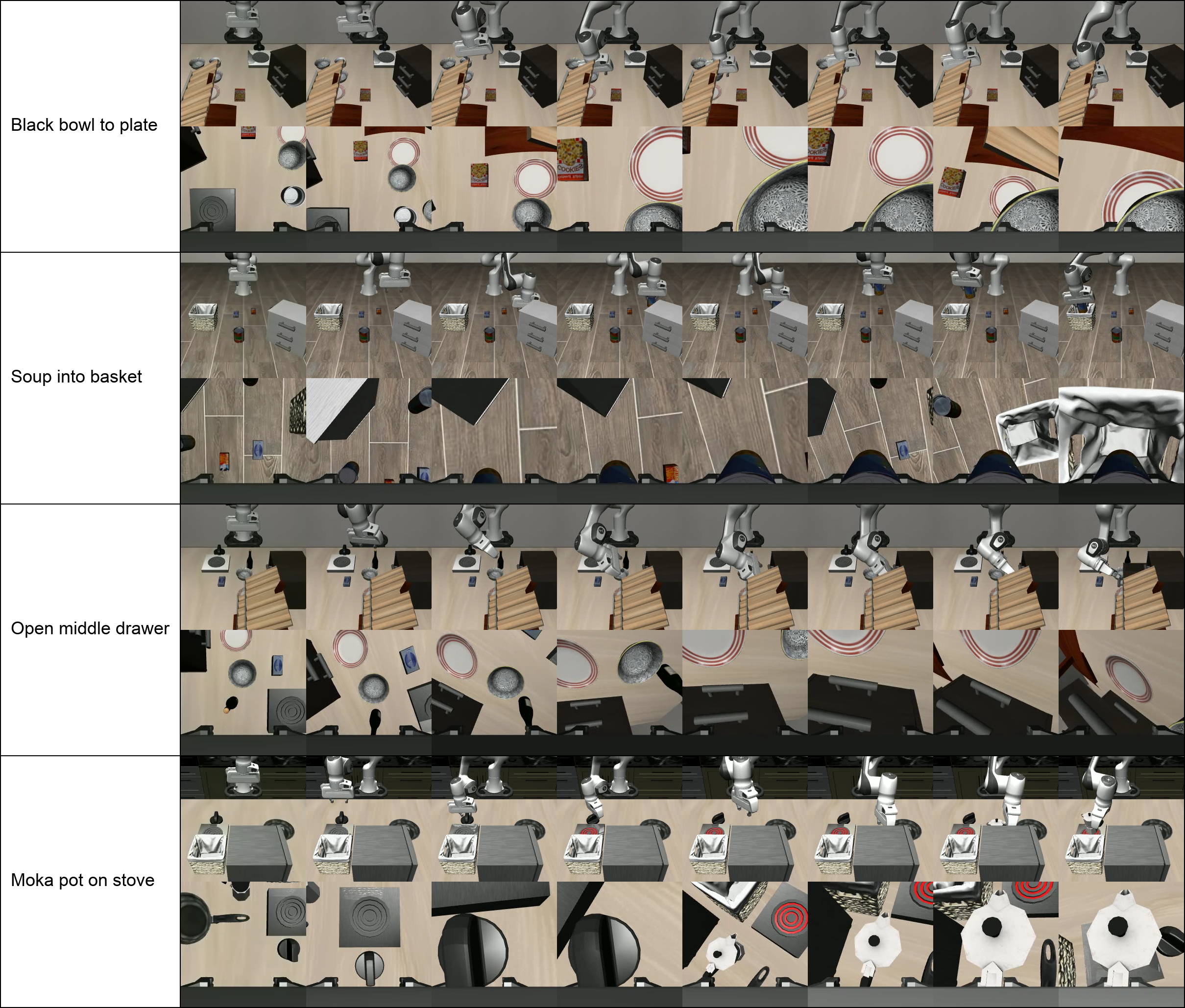}
\caption{Example rollouts on LIBERO-Occ.}
\label{fig:liberoocc_rollouts}
\end{figure}

LIBERO-Occ evaluates robustness to scene-induced occlusion across the Spatial, Object, Goal, and Long suites. All official benchmark initial states are evaluated, corresponding to 500 rollouts for each suite and 2,000 rollouts in total. Success is determined by the environment-defined task success criterion.Example rollouts are shown in Fig.~\ref{fig:liberoocc_rollouts}

\subsubsection{RLBench-OG Occlusion1}
\label{app:rlbench}

\begin{figure}[h]
\centering
\includegraphics[width=\linewidth]{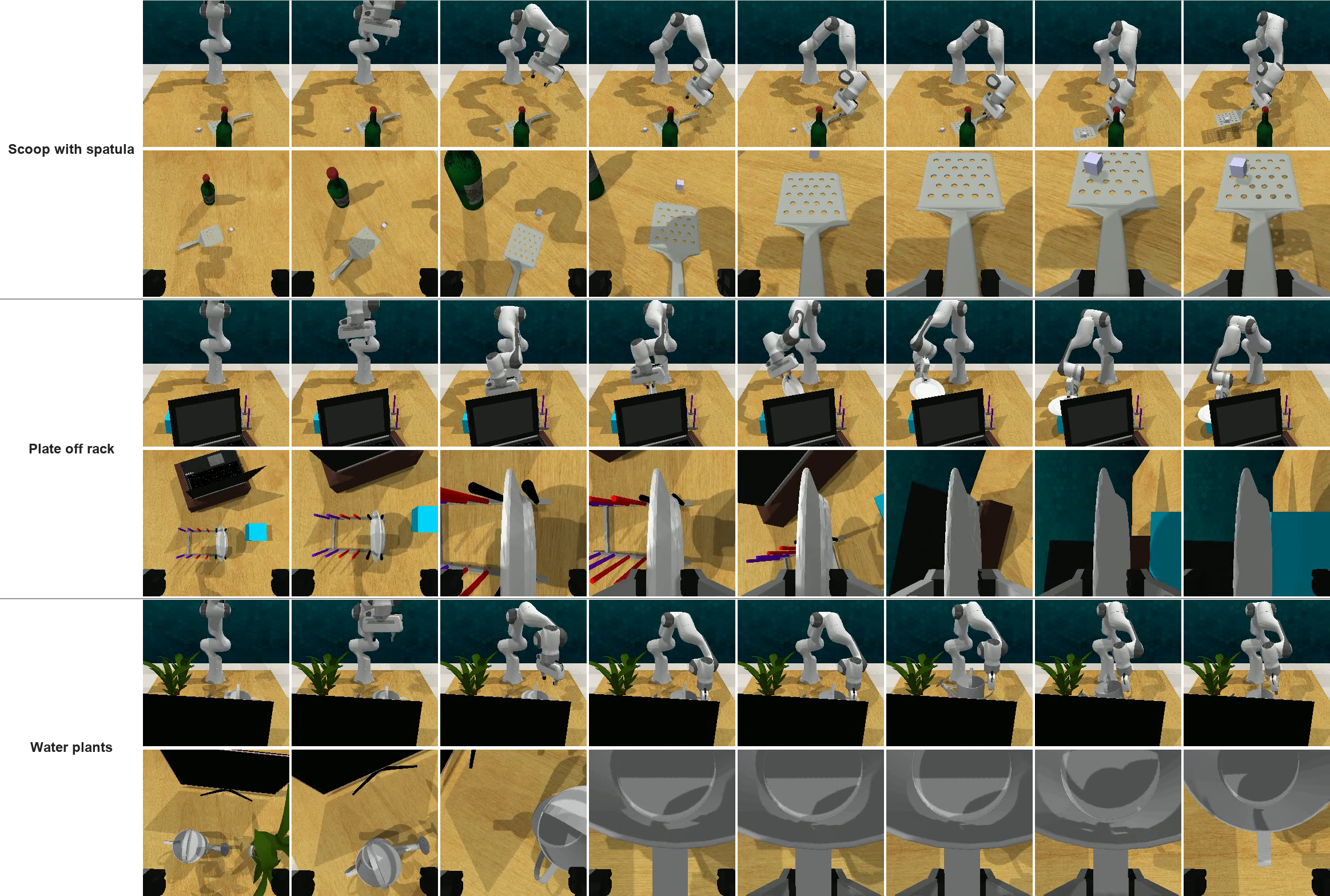}
\caption{Example rollouts on RLBench-OG Occlusion1.}
\label{fig:rlbench}
\end{figure}

Evaluation uses the ten tasks of the RLBench-OG Occlusion1 setting: basketball in hoop, scoop with spatula, take plate off colored dish rack, water plants, block pyramid, solve puzzle, take USB out of computer, close drawer, straighten rope, and toilet seat down. Example rollouts are shown in Fig.~\ref{fig:rlbench}.

For RLBench-OG Occlusion1, models are trained and evaluated on the occluded task configurations following the benchmark protocol. RLBench-OG uses a control interface different from LIBERO. The same \method{} mechanism---trajectory-conditioned predictive perception, memory construction, and memory-conditioned closed-loop action generation---is retained with the benchmark-specific controller.

\subsubsection{Real-World Task Details}

\label{app:real_world_tasks}

\begin{figure}[t]
  \centering
  \includegraphics[width=\linewidth]{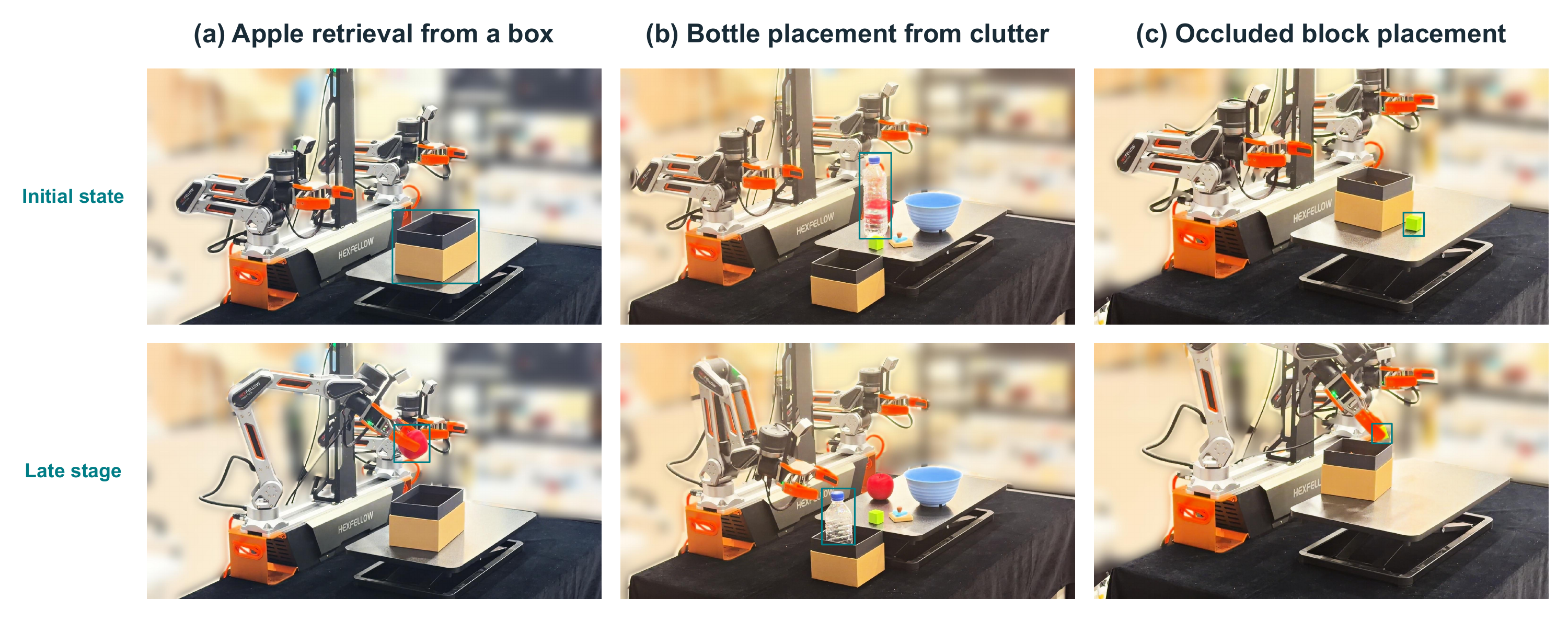}
  \caption{Third-person views (for record only) of real-robot inference on three real-world tasks, with one early and one late frame per task.}
  \label{fig:third_person_rollouts}
\end{figure}

Three manipulation tasks are evaluated on an ARCHER-Y6 arm with a GP-100 gripper and scene-level and wrist-mounted Intel RealSense D435 cameras. Each task is evaluated over 30 trials. \method{} constructs its predictive world memory once from the initial observations and retrieves this context alongside live RGB observations and proprioception during closed-loop control. The predicted camera trajectory specifies virtual viewpoints; it is not executed as a separate physical inspection motion. The third-person views in Figure~\ref{fig:third_person_rollouts} illustrate early and late execution stages. Descriptions of initial target occlusion below refer to the robot-camera observations.

\paragraph{Retrieving an apple from a box.}
A red apple is placed inside an open box, whose walls partially obscure the apple in the initial robot observations. The challenge is to infer the grasp region and its relation to the surrounding walls from incomplete visual evidence. Task-conditioned virtual view planning is designed to provide predictive context for the partially observed box interior, while geometry-conditioned world memory retains these spatial features for action generation. Live observations support gripper positioning and extraction as the scene changes. The selected late frame shows the apple held above the rim, making the retrieval outcome visible from the external viewpoint.

\paragraph{Picking a bottle from clutter and placing it in a box.}
A plastic bottle stands on a tabletop among a bowl, an apple, and smaller objects, with an open receiving box nearby. The robot should grasp the bottle, lift it from the clutter, transfer it to the box, and release it inside. This task combines target selection in a visually cluttered scene with coordination between a grasp location and a separate placement region. Also, half of the bottle lies outside the initial robot view, thus complicating the interpretation of the bottle's local geometry and the gripper approach. \method{}'s task-conditioned virtual-view predictions are intended to complement the initial observations with spatial context, which the world memory retains throughout the transfer. Memory retrieval complements current observations, while closed-loop action prediction follows the changing bottle and gripper states. The late frame shows the bottle standing inside the receiving box as the gripper withdraws.

\paragraph{Grasping an occluded block and placing it in a box.}
A green block is placed on the tabletop beside a box, which initially occludes the target in the robot-camera observations. The robot approaches the block, closes the gripper around it, lifts it above the box opening, and deposits it inside. The box therefore serves both as an occluder and as the destination, requiring the policy to relate the initially hidden grasp region to the visible container geometry. The cached memory remains available as live observations reveal the target and guide the subsequent grasp and transfer. The selected late frame captures the block held above the opening before release; completion additionally requires depositing the block inside the box.

\subsection{Limitations}
\label{app:limitations}

The current system relies on calibrated cameras and depth information to construct geometry-aware virtual observations, which introduces sensing assumptions beyond purely image-based policies. Besides, the current system constructs $M_0$ once from the initial scene and keeps it fixed throughout manipulation. When objects are substantially displaced or articulated, parts of the cached memory may become stale. Fresh live observations provide the primary control feedback, but the memory itself is not explicitly updated. Dynamically scheduled memory reconstruction is a promising direction for long-horizon tasks.

\end{document}